\documentclass[letterpaper, 10 pt, conference]{ieeeconf}  
\usepackage{amsmath,amsfonts}
\usepackage{algorithmic}
\usepackage{algorithm}
\usepackage{array}

\usepackage[caption=false,font=normalsize,labelfont=sf,textfont=sf]{subfig}
\usepackage{textcomp}
\usepackage{stfloats}
\usepackage{url}
\usepackage{verbatim}
\usepackage{graphicx}
\usepackage{bbm}
\usepackage{bbold}
\usepackage{cite}
\usepackage{float}
\usepackage{booktabs}
\usepackage{setspace}
\usepackage{multirow}
\usepackage[utf8]{inputenc}
\usepackage[T1]{fontenc}
\usepackage{geometry} 

\IEEEoverridecommandlockouts                              
\title{\LARGE \bf
Self-Supervised Traversability Learning with Online Prototype Adaptation for Off-Road Autonomous Driving
}

\author{Yafeng Bu, Zhenping Sun, Xiaohui Li, Jun Zeng, Xin Zhang and Hui Shen
\thanks{\textit{(Yafeng Bu and Zhenping Sun are co-first authors.)} \textit{(Corresponding author: Hui Shen.)}}
\thanks{The authors are with the College of Intelligence Science and Technology,         National University of Defense Technology,410073 Changsha,China (email:shenhui@nudt.edu.cn)}%
}
\begin{document}

\maketitle
\thispagestyle{empty}
\pagestyle{empty}

\begin{abstract}


Achieving reliable and safe autonomous driving in off-road environments requires accurate and efficient terrain traversability analysis. However, this task faces several challenges, including the scarcity of large-scale datasets tailored for off-road scenarios, the high cost and potential errors of manual annotation, the stringent real-time requirements of motion planning, and the limited computational power of onboard units. To address these challenges, this paper proposes a novel traversability learning method that leverages self-supervised learning, eliminating the need for manual annotation. For the first time, a Bird’s-Eye View (BEV) representation is used as input, reducing computational burden and improving adaptability to downstream motion planning. During vehicle operation, the proposed method conducts online analysis of traversed regions and dynamically updates prototypes to adaptively assess the traversability of the current environment, effectively handling dynamic scene changes. We evaluate our approach against state-of-the-art benchmarks on both public datasets and our own dataset, covering diverse seasons and geographical locations. Experimental results demonstrate that our method significantly outperforms recent approaches. Additionally, real-world vehicle experiments show that our method operates at 10 Hz, meeting real-time requirements, while a 5.5 km autonomous driving experiment further validates the generated traversability cost map’s compatibility with downstream motion planning.
\end{abstract}
\section{INTRODUCTION}
\label{INTRODUCTION}
 Terrain traversability analysis is an important ability for autonomous vehicles to navigate and understand the environment in complex off-road environments. However, compared with on-road scenarios \cite{city1,city2,city5}, the traversability of off-road environments is more ambiguous and lacks standards, making it difficult to simply classify them as traversable or untraversable. Specifically, factors such as the vehicle's interaction with the terrain and surface materials affect traversability, making it more difficult to determine traversal costs for different terrain based on human experience \cite{jung2024v}. Although certain efforts have been made in datasets for off-road environments \cite{dataset4,dataset6,dataset7}, these datasets are still limited to specific geographical and seasonal conditions, and the cost of collecting and manually annotating various outdoor scenes is prohibitively high.
 
Due to the aforementioned limitations, supervised learning for traversability prediction faces numerous challenges, particularly when relying on large amounts of precisely labeled data. In \cite{UFO}, to avoid dependence on fine-grained annotations, a pre-trained model is used to generate pseudo-labels for the data. However, the class prediction results of such methods are difficult to directly map to traversability metrics or traversal costs, as there is no clear correspondence between semantic categories and traversability.

\begin{figure}[!t]
	\centering
	\subfloat{
		\centering
		\includegraphics[width = 0.98\linewidth]{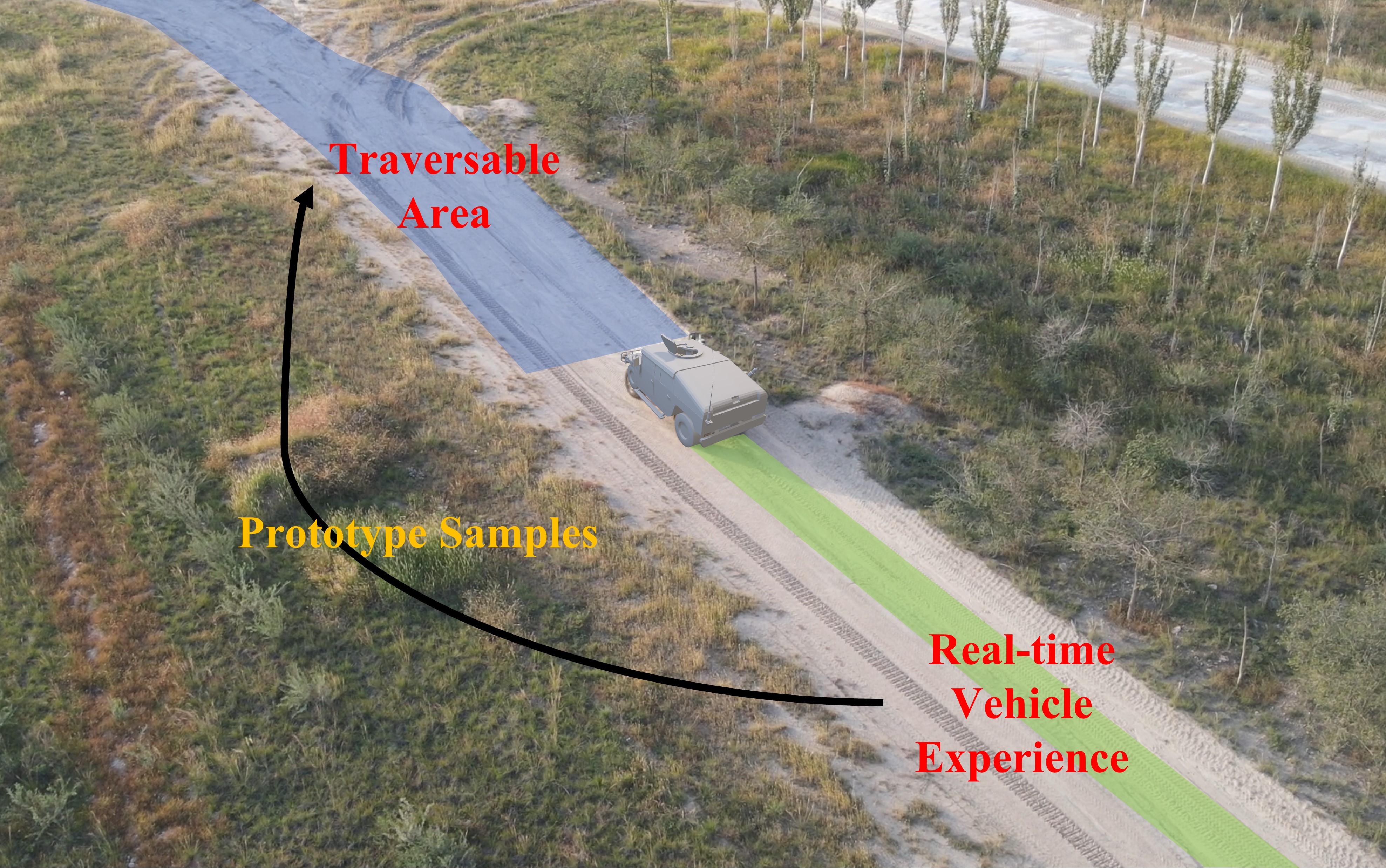}
	}
	\vspace{1mm} 
	\subfloat{
		\centering
		\includegraphics[width = 0.448\linewidth]{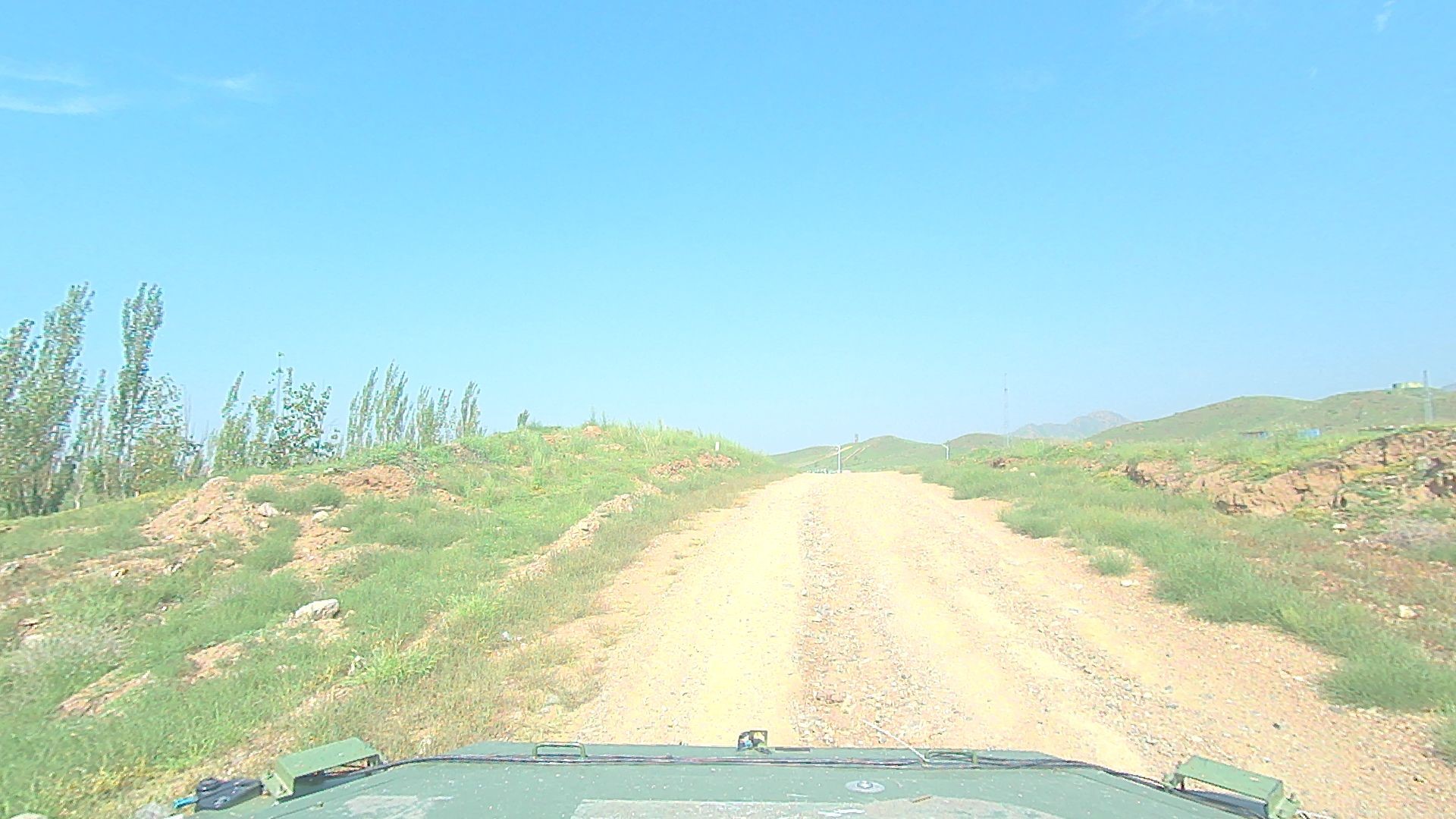}
		\includegraphics[width = 0.253\linewidth]{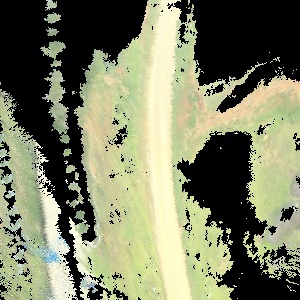}
		\includegraphics[width = 0.253\linewidth]{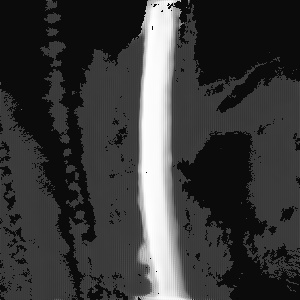}
	}
	\caption{(Top) Building an environmental hypothesis model based on real-time vehicle experience: Predicting the traversability of unexplored regions by extracting prototype vectors from traveled areas. (Bottom-left) RGB images captured by the onboard front-facing camera. (Bottom-middle) BEV generated by integrating LiDAR point clouds, RGB images, and odometry data. (Bottom-right) Traversability analysis results based on real-time experience.}
	\label{fig:label}
\end{figure}

To address this issue, self-supervised learning has been demonstrated as a promising research direction for traversability prediction in off-road environments  \cite{self-supervised1,self-supervised3,self-supervised5,self-supervised6,self-supervised7,jung2024v,self-supervised9,seo2023learning,self-supervised11,schmid2022self}. These methods typically utilize human driving data for automatic labeling without requiring manual intervention. However, existing self-supervised methods still have certain limitations: since labeled samples typically cover only traversable areas, a large portion of environmental observations remain unlabeled, leading to biases in model training. Self-supervised methods that train on a single predictive factor are prone to overfitting \cite{schmid2022self}. Existing methods commonly use front-view images as input. Although front-view images provide rich semantic and geometric information \cite{jung2024v}, they also contain numerous background elements unrelated to traversability (e.g., the sky), leading to wasted computational resources. Additionally, the nature of front-view images makes them difficult to directly apply to downstream motion planning tasks.

In summary, traversability analysis in off-road environments faces several key challenges: First, the lack of precisely labeled datasets, the high cost of manual annotation, and the difficulty of achieving comprehensive labeling make self-supervised methods that rely solely on human driving data for single-class labeling prone to overfitting. Second, the performance and power constraints of onboard computing units, along with high-resolution front-view images containing numerous irrelevant elements, increase computational burden and make real-time processing challenging. Furthermore, front-view inputs are not conducive to motion planning tasks. Lastly, the complexity and variability of off-road environments make it difficult for offline-trained models to adapt to dynamic scene changes.

To address the aforementioned challenges, we propose a novel self-supervised traversability estimation method based on BEV. We introduce an automatic annotation process that generates reliable self-supervised labels on BEV by leveraging historical vehicle trajectories and reliable obstacle detection results. Under these labels, contrastive representation learning and clustering methods are employed to learn traversability representations for complexly distributed off-road environments. Furthermore, to compensate for limited offline training datasets that fail to cover all traversable sample distributions, we integrate online clustering methods to achieve adaptive real-time environmental traversability analysis. 

To validate our method's effectiveness, we collected off-road scenario data covering seasonal variations, diverse locations, and varying illumination conditions. Experimental results demonstrate that our method effectively learns traversability in diverse unstructured environments. Finally, we demonstrate motion planning applications based on the traversability analysis results, proving its superior suitability for downstream motion planning tasks. Our contributions are as follows:
\begin{itemize}
	\item We propose an automatic labeling method that generates reliable self-supervised labels using high-confidence obstacle information detected by LiDAR and vehicle trajectories. Additionally, we leverage clustering algorithms to explore unlabeled samples, enhancing the model's representational capability.
	\item BEV-based input representation, which is more compatible with downstream motion planning and effectively filters out irrelevant visual elements (e.g., sky) typically present in front-view images, thereby reducing computational load.
	\item We implement adaptive real-time traversability analysis by dynamically analyzing traversed areas during autonomous driving and updating prototypes online. Similarity is used as a traversability metric to construct a probabilistic environmental representation, enhancing the model's adaptability to dynamic scene changes.
\end{itemize}
\section{Related Works}
\label{Related Works}
\subsection{Traditional Traversability Analysis}
Many early studies evaluated the traversability of the environment by analyzing geometric or visual features \cite{early1,early2,early3,early4,early5}. With the successful application of deep learning in semantic segmentation, a large number of learning-based approaches for terrain traversability analysis have emerged \cite{ref6}. Suryamurthy et al. \cite{ref7} used SegNet and ERFNet to predict pixel-wise labels of terrain. Guan et al. \cite{ref8} use fully convolutional neural networks to identify safe and travelsable regions in off-road environments from RGB images.  Sun et al. \cite{ref9} improve PSPNet and achieves better performance in traversability detection in off-road environments. Kim et al.\cite{UFO} use LiDAR image fusion for accurate semantic on the bird’s-eye view, improving the accuracy of the generated semantic map.  However, these methods heavily rely on training data and require extensive manual annotation.  The generalization capability of supervised learning is limited, and when faced with data that is not included in the training data distribution, it may provide incorrect estimations. Similarly, in complex off-road environments, traversability does not strongly correlate with semantic categories. The same semantic category may correspond to different traversability \cite{ref10}, making it difficult to assign reasonable traversability costs to the environment based on semantic categories.

\subsection{Self-Supervised Learning of Traversability}
Given the dependency of supervised learning on data annotation, self-supervised learning methods have gained significant attention, avoiding the high cost of manual annotation and utilizing the driving experience of vehicles to learn terrain traversability. These approaches acquire knowledge of environmental traversability via self-annotation. For instance, Castro et al.\cite{self-supervised11} utilized IMU z-axis measurements as traversability scores, whereas Wellhausen et al.\cite{wellhausen2019should} applied mechanical sensors to assess the mechanical feedback from robot-terrain interaction, adopting it as a criterion for traversability analysis. Schmid et al.\cite{schmid2022self} proposed an anomaly detection method based on an autoencoder, which calculates the loss only for trajectories successfully traversed by the vehicle, and evaluates terrain traversability through reconstruction error during inference. Nonetheless, the excessive generalization of latent representations in autoencoders can cause false negatives\cite{gong2019memorizing}, potentially assigning incorrectly low traversal costs to non-traversable regions due to small reconstruction errors, thereby directly affecting the safety of path planning in autonomous navigation systems.

\subsection{Contrastive Traversability Learning}
Contrastive learning, by constructing positive and negative sample pairs, can effectively learn terrain feature representations, providing a discriminative feature space for traversability prediction. Xue et al.\cite{self-supervised1} proposed the use of prototype vectors, learning from positive samples and unlabeled samples to create pseudo-labels. Seo et al.\cite{seo2023learning} utilized the trajectories traversed by vehicles as positive samples, employing Positive-Unlabeling (PU) learning and a normalizing-flow model to learn traversability in complex off-road environments. Moreover, Seo et al.\cite{seo2023learning} used contrastive loss on augmented images to promote effective image representations in the model. However, this contrastive loss may not be sufficient to provide meaningful information in distinguishing between traversable and non-traversable areas. 

To enable the model to extract features containing more meaningful information, Jung\cite{jung2024v} employed contrastive learning, using traversed trajectories as positive samples and sampling areas outside the positive samples as negative samples, and used SAM's mask prediction as an additional signal for traversability to mitigate the inclusion of positive samples in the sampled negative samples. However, in scenarios with weak discriminability, SAM might make incorrect predictions, adversely affecting the training. In contrast, we employ a highly robust obstacle detection algorithm\cite{Xue2023} to identify negative samples in the environment, preventing the negative impact on training from erroneous negative sample selection.
\section{METHOD}
\label{METHOD}
Our objective is to establish a mapping from BEV to pixel-level feature maps, with online updating of traversable prototype features. The traversability probability is characterized by computing a similarity metric between the feature maps and prototypes. In this section, we first describe the BEV generation process, followed by the self-supervised label generation and training procedure. Finally, we elaborate on prototype updating and the conversion to traversability metrics. The overview of our method is illustrated in Fig. \ref{Train}.
\subsection{Generating environmental Bird’s-Eye View}
We employ an environmental BEV for traversability analysis. The BEV projects the surrounding environment into a top-down representation, effectively filtering out irrelevant information (such as the sky), which is more suitable for downstream motion planning task. The BEV construction pipeline is shown in Fig. \ref{GenerateBEV}. First, we use calibrated extrinsic parameters $(R_{\text{L2C}},T_{\text{L2C}})$ and camera intrinsics $K$, we project LiDAR point $(x,y,z)$ to pixel coordinates $(u,v)$ via:
\begin{equation}
	\left( \begin{array}{l}
		u\\
		v\\
		1\\
	\end{array} \right) =K\left( R_{\text{L2C}}\mid T_{\text{L2C}} \right) \left( \begin{array}{l}
		x\\
		y\\
		z\\
		1\\
	\end{array} \right) .
\end{equation}

Since LiDAR point clouds are relatively sparse, the amount of information contained in a single frame is limited. Therefore, we fuse multiple frames of point clouds using the LiDAR odometry. First, at time $t$, we transform the fused point cloud from time $t-1$ into the vehicle's coordinate system at time $t$. The process is as follows:
\begin{equation}
	P_{\text{veh}}\left( t \right) =\left( R_{t}^{T}\cdot P_{\text{Odom}}\left( t-1 \right) +\left( -R_{t}^{T}\cdot T_t \right) \right) \cup P_t,
\end{equation}
Here, $P_{\text{veh}}(t)$ represents the fused point cloud at time $t$ in the vehicle's coordinate system, while $R_t$ and $T_t$ denote the results from the LiDAR odometry at time $t$. $P_{\text{Odom}}(t-1)$ refers to the fused point cloud at time $t-1$ in the odometry coordinate system. $P_t$ represents the point cloud observed by the sensor at time $t$. After computing the fused point cloud at time $t$, it is transformed into a BEV, resulting in the environmental BEV $M_t$. For the fused point cloud $P_{\text{veh}}(t)$ at time $t$, we transform it into the odometry coordinate system using the odometry results at that time, in preparation for the fusion at the next time step:
\begin{figure}[!t]
	\centering
	\includegraphics[width=3in]{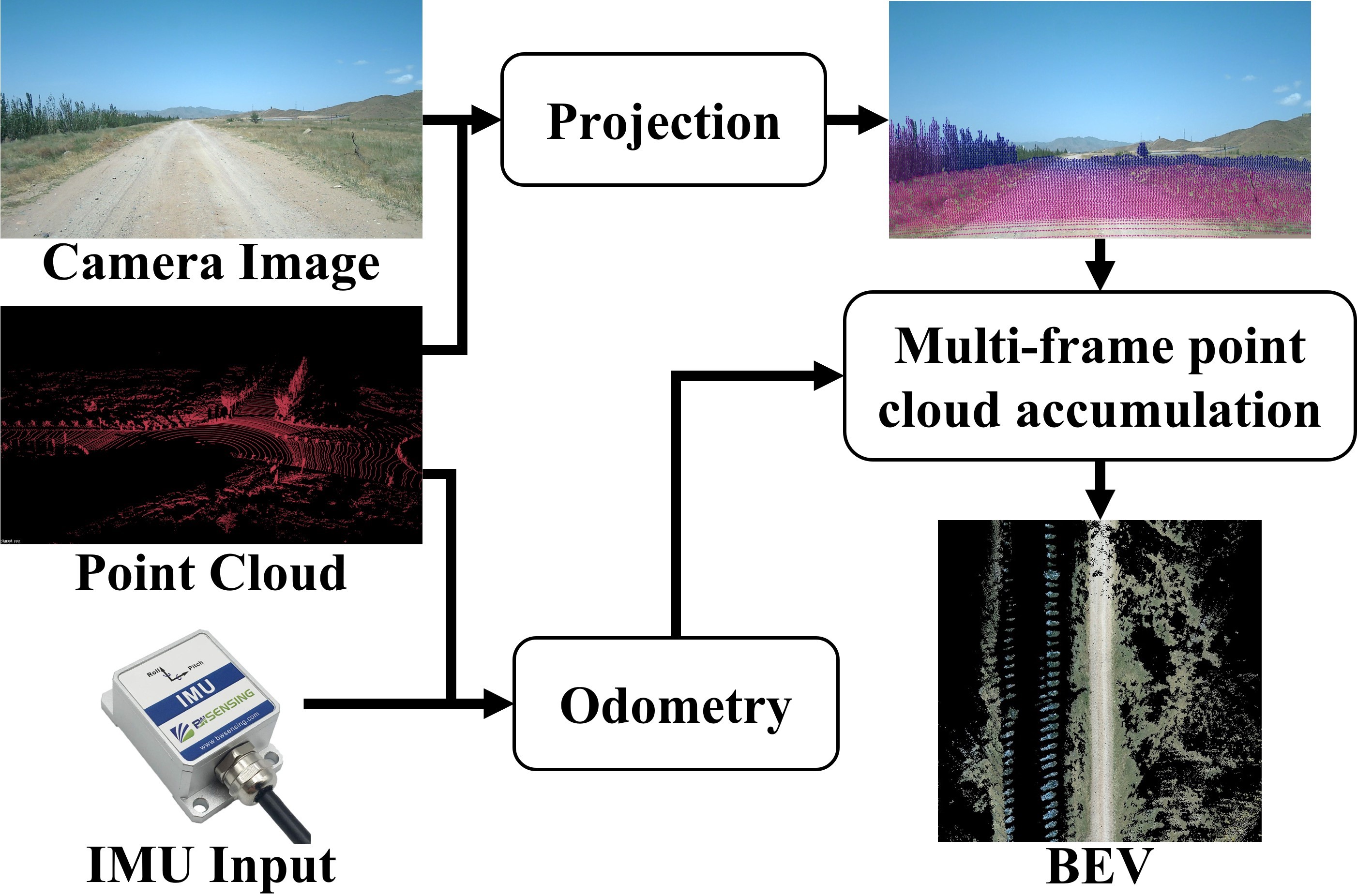}
	\caption{The BEV generation process: For each frame of temporally synchronized point clouds and RGB images, colorize the point clouds using a projection matrix, and accumulate multiple frames of point clouds relying on odometry to complete the BEV generation.}
	\label{GenerateBEV}
\end{figure}
\begin{equation}
	P_{\text{Odom}}\left( t \right) =P_{\text{VehicleCoor}}\left( t \right) \cdot R_t+T_t.
\end{equation}
\subsection{Generating Self-supervised Label}
The self-supervised labels are generated through an automatic process from the BEV collected during driving. During the data collection process, we record the regions traversed by the vehicle as well as the LiDAR perception results. Regions traversed by the vehicle are considered safe, while areas confidently identified as obstacles by the perception module\cite{Xue2023} are classified as non-traversable. These annotations serve as self-supervised signals for learning environmental traversability. Specifically, for any point on the vehicle's trajectory at a given time, we define the set of contact points between the four wheels of the vehicle and the ground as $J = \left\{ J_{lf}, J_{lr}, J_{rf}, J_{rr} \right\}$, where $J_{lf}$, $J_{lr}$, $J_{rf}$, and $J_{rr}$ represent the contact positions of the left front wheel, left rear wheel, right front wheel, and right rear wheel with the ground, all expressed in the vehicle's coordinate system. For any trajectory point set $J$ at a past or future time $\tau$, the coordinates in the vehicle's coordinate system at the current time $t$ are obtained through coordinate transformation using the odometry results:
\begin{equation}
	\label{coordinate transformation}
	J^{\tau ,t}=\left( [R|T]^t \right) ^{-1}\cdot [R|T]^{\tau}\cdot J,
\end{equation}                                 

Here, $J^{\tau,t}$ represents the set of vehicle trajectory points after coordinate transformation, while $[R|T]^t$ and $[R|T]^\tau$ denote the pose transformation matrices corresponding to times $t$ and $\tau$, respectively. Using \eqref{coordinate transformation}, the vehicle's trajectory points from both past and future time intervals can be consistently represented in the vehicle's coordinate system at the current time $t$. Subsequently, these trajectory points are further projected onto the $M_t$, enabling automatic labeling of traversable areas. 

Since both the BEV $M_t$ and the LiDAR-based obstacle detection result $O_t$ are computed from multi-frame point clouds and are synchronized in both time and space, we can easily use the obstacle detection result $O_t$ to automatically label the non-traversable areas in $M_t$.
\begin{figure*}[!t]
	\centering
	\includegraphics[width=1\textwidth]{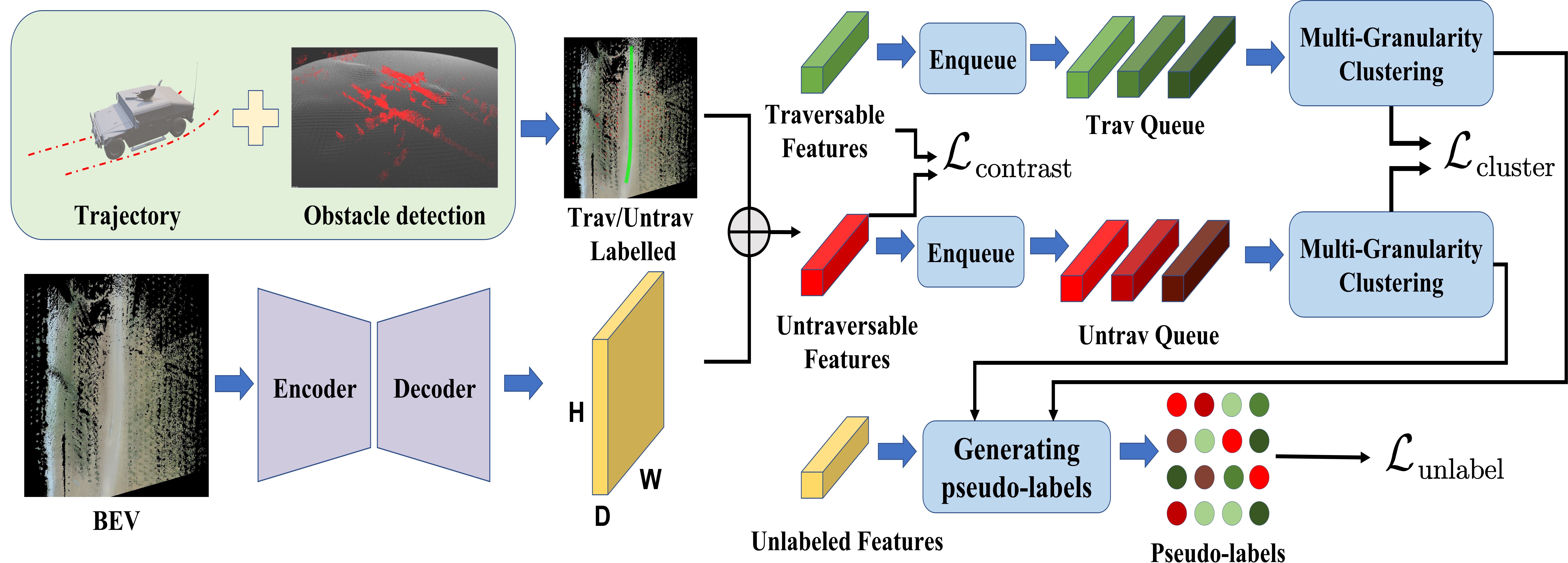}
	\caption{Overview of training process. First, we map the vehicle trajectory and obstacle detection results into the BEV space to generate self-supervised labels for each frame of the BEV image. A ResNet-based encoder-decoder architecture is designed, where the encoder extracts multi-level features, and the decoder generates traversability feature maps. Then, we extract traversable and untraversable features, as well as features from unlabeled samples, and assign pseudo-labels to the unlabeled samples through clustering. Finally, by optimizing the combined loss function, the network parameters are updated using the backpropagation algorithm.}
	\label{Train}
\end{figure*}
\subsection{Learning Traversability}
In this section, we propose a method for learning environmental traversability features using only self-supervised labels. For the BEV $M_t \in \mathbb{R}^{W \times H \times 3}$ at time $t$, we define a traversability feature extraction network $f(M_t) = h \circ g(M_t)$ to predict the traversability feature map $F_t \in \mathbb{R}^{W \times H \times D}$, where $D$ represents the dimensionality of the traversability features. The model $f(x)$ consists of an encoder $g(\cdot)$ and a decoder $h(\cdot)$. Through this model, we obtain a feature map of the same size as the original image. The core objective of the training is to map the raw input to an embedding space through the model, such that samples from different classes are well-separated, i.e., embeddings from the same class are close to each other, while embeddings from different classes are distant. Specifically, for each frame of the BEV $M_i$ in the training samples, there is a corresponding auto-labeled result $L_i$. We use the constructed model $f(x)$ to compute the feature map $F_i$, which consists of three sets: the traversable sample feature set $F_{\text{trav}}$, the non-traversable sample feature set $F_{\text{untrav}}$, and the unlabeled sample feature set $F_{\text{unlabel}}$. Multiple optimization objectives are designed to compute the loss, thereby optimizing the network parameters.The overall architecture of our learning method is illustrated in Fig. \ref{Train}.

For each sample feature $z_i$ in the traversable sample feature set $F_{\text{trav}}$, we aim to minimize the distance between $z_i$ and the other traversable sample features $z_j \in F_{\text{trav}}$ ($j \ne i$), while maximizing the distance between $z_i$ and each feature sample $z_k \in F_{\text{untrav}}$ in the non-traversable sample feature set $F_{\text{untrav}}$. To achieve this goal, the $\text{InfoNCE}$ loss is used to calculate, as shown in eq.\eqref{InfoNCE}:
\begin{equation}
	\label{InfoNCE}
	\begin{array}{l}
		\mathcal{L}_{\text{contrast}}(F_\text{trav},F_\text{untrav})=\\
		-\frac{1}{N(N-1)} \mathop{\sum}\limits_{z_{i} \in F_{\text{trav}}} \mathop{\sum}\limits_{z_{j} \in F_{\text{trav}}} \mathbb{1}(i \neq j) \log \frac{\exp \left(z_{i}^{T} \cdot z_{j} / \tau\right)}{\mathop{\sum}\limits_{z_{k} \in F_{\text{untrav}}} \exp \left(z_{i}^{T} \cdot z_{k} / \tau\right)}
	\end{array}.
\end{equation} 
where $N=|F_{\text{trav}}|$ and $\tau$ denotes a temperature scalar. In real-world scenarios, there are multiple categories of traversable and non-traversable samples. To address this, we introduce the concept of prototypes \cite{PSL, PCL, PICO}. In this task, we define a prototype as the centroid of a cluster formed by similar samples and assign each sample to multiple prototypes of different granularities to ensure that each sample embedding is closer to its relevant prototype. For both traversable and untraversable sample sets, we construct prototype hierarchies through multi-scale clustering. Let $Q_{\text{trav}}\in \mathbb{R}^{N\times D}$ and  $Q_{\text{untrav}}\in \mathbb{R}^{N\times D}$ denote the feature queues containing all training samples from traversable and untraversable classes respectively.  We perform $M$ independent K-Means clusterings on each queue with progressively increasing cluster numbers $K=\{k_m\}_{m=1}^M$. This produces hierarchical prototype sets:
\begin{equation}
	\begin{aligned}
			\mathcal{P}_\text{trav} &= \{ \{p_{m,i}^\text{trav}\}_{i=1}^{k_m} \}_{m=1}^M \\\mathcal{P}_\text{untrav} &= \{ \{p_{m,i}^\text{untrav}\}_{i=1}^{k_m} \}_{m=1}^M
		\end{aligned},
\end{equation}
where $p_{m,i}^\text{trav}$ and $p_{m,i}^\text{untrav}$ represent the $i_{\text{th}}$ prototype in the $m_{\text{th}}$ clustering group for traversable and untraversable classes respectively. Each sample is assigned to its nearest prototype at multiple granularities, enabling hierarchical feature alignment through prototype-guided contrastive learning.
For the feature set $F$ and the prototype set $\mathcal{P}$, we use the same method of Li et al.\cite{PCL}, which randomly sampling $r$ negative prototype samples to compute the normalization term.
\begin{equation}
	\mathcal{L}_{\text{Proto}}(F,\mathcal{P})=-\frac{1}{N M}\sum_{i=1}^{N}\sum_{m=1}^{M} \log \frac{\exp \left(z_{i} \cdot p_{s,m}^{\text{pos}}/ \tau\right)}{\sum_{j=0}^{r} \exp \left(z_{i} \cdot p_{j,m}^{\text{neg}} / \tau\right)},
\end{equation}
Here, $z_i \in F$, $p_{s,m}^{\text{pos}}$ is the prototype sample corresponding to $z_i$ in the $m_{\text{th}}$ cluster, $p_{j,m}^{\text{neg}}$ is the negative prototype sample corresponding to $z_i$ in the $m_{\text{th}}$ cluster, and $\tau$ is the temperature coefficient. Therefore, given $z_i \in F_{\text{trav}}$ and a set of prototype samples $p_m = \{p^{\text{trav}}_m \cup p^{\text{untrav}}_m \}$, we only reduce the distance between $z_i$ and its corresponding prototype sample $p_{s,m}^{\text{trav}} \in p^{\text{trav}}_m$, while increasing the distance between $z_i$ and the $r$ negative prototype samples sampled from $p^{\text{untrav}}_m$. The same applies to the traversability feature set $F_{\text{untrav}}$. Thus, the clustering loss $\mathcal{L}_{\text{cluster}}$ can be defined as:
\begin{equation}
	\mathcal{L}_{\text{cluster}}=\mathcal{L}_{\text{Proto}}(F_\text{trav},\mathcal{P})+\mathcal{L}_{\text{Proto}}(F_\text{untrav},\mathcal{P}),
\end{equation}
Here, $\mathcal{P}=\{\mathcal{P}^{\text{trav}}\cup \mathcal{P}^{\text{untrav}}\}$. 
For any unlabeled sample $z_i^\text{unlabel} \in F_\text{unlabel}$, we compute the corresponding prototype sample $p_{s,m}^{\text{unlabel}}$ based on the union of the traversable and non-traversable prototype sample sets of the $m_{\text{th}}$ group, $\mathcal{P}_m = \mathcal{P}_m^\text{trav} \cup \mathcal{P}_m^\text{untrav}$, as follows:
\begin{equation}
	\label{unlabelprototype}
	p^{\text{unlabel}}_{s,m}={\arg\max}_{p_{i,m}\in \mathcal{P}_m}(\frac{{z_{i}^{\text{unlabel}}}^T \cdot p_{i,m}}{\|{z_{i}^{\text{unlabel}}}\|\|p_{i,m}\|}).
\end{equation}
To improve within-cluster compactness for conventional instance alignment while avoiding class collision issue, we adopt the Positive Sample Alignment (PSA) loss\cite{PSL}, sampling neighboring samples through a Gaussian distribution:
\begin{equation}
	v_i=z_i+\sigma \boldsymbol{\epsilon}, \quad \text { where } \quad {\epsilon} \sim \mathcal{N}(0, {I}) ,
\end{equation}
Here, $\sigma$ represents a positive hyperparameter. Then we use the PSA loss to compute the clustering loss for each group on the unlabeled sample feature set $F_\text{unlabel}$, the loss $\mathcal{L}_{\text{unlabel}}$ for unlabeled samples set is can be defined as:
\begin{equation}
	\label{prototype}
	\begin{aligned}
		\mathcal{L}_{\text{unlabel}} & =\frac 1 {N} \frac 1 {M}\sum_{i=1}^N\sum_{m=1}^M\left\|v_i-p_{s,m}^\text{unlabel}\right\|_{2}^{2} \\
		& =\frac 1 {N}\sum_{i=0}^N\left\|(z_i+{\sigma} {\epsilon})-p_{s,m}^\text{unlabel}\right\|_{2}^{2}
	\end{aligned},
\end{equation}
Here, $z_i\in F_\text{unlabel}$ and $p_{s,m}^\text{unlabel}$ is the corresponding prototype sample of $z_i$. Finally, our final loss is represented as:
\begin{equation}
	\mathcal{L}=\mathcal{L}_{\text{contrast}}+ \lambda * (\mathcal{L}_{\text{cluster}}+\mathcal{L}_{\text{unlabel}}),
\end{equation}
with weight $\lambda\in [0,1]$. 
\subsection{Computing Traversability Cost}
\begin{figure}[!t]
	\centering
	\includegraphics[width=1\linewidth]{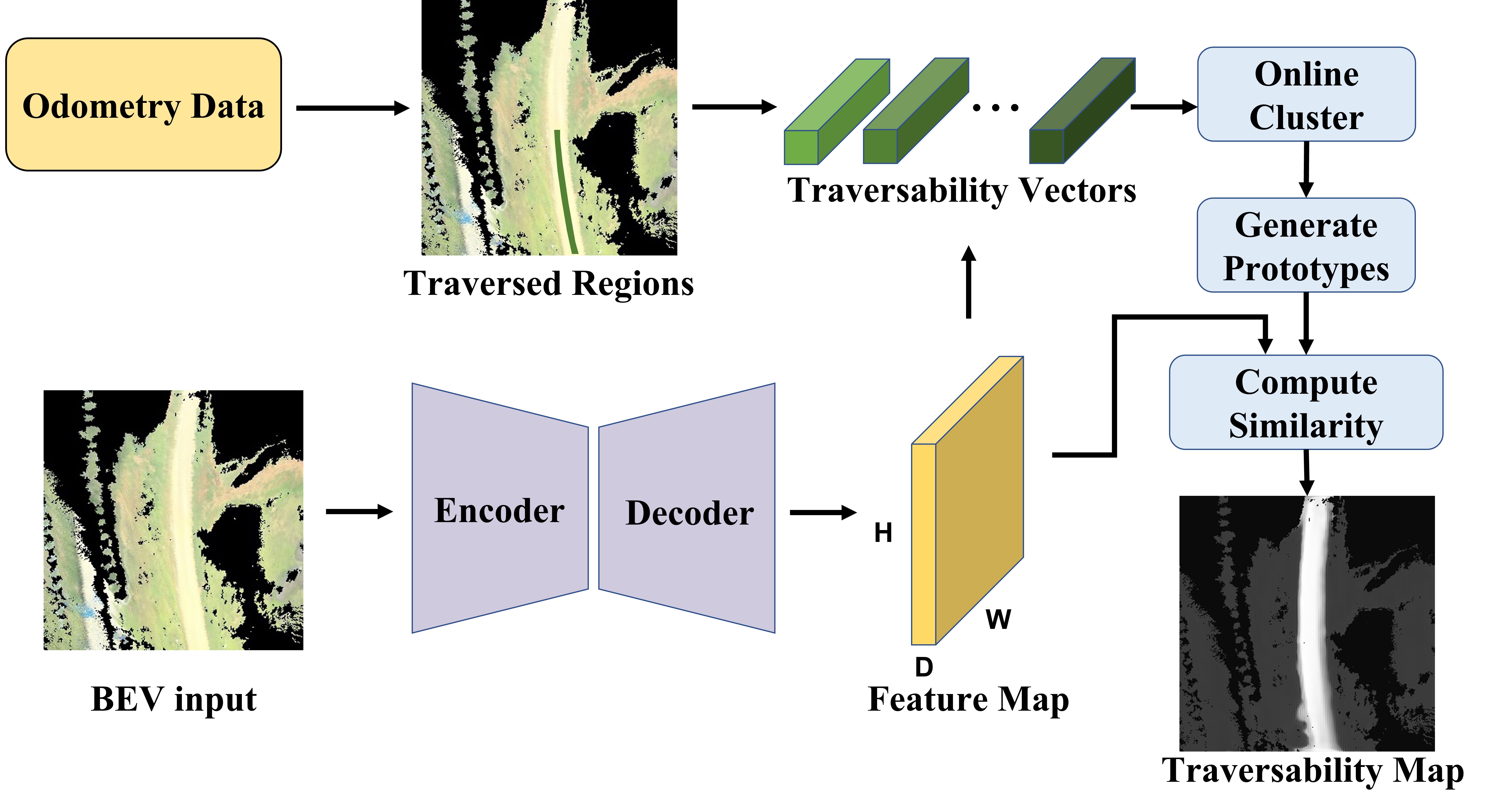}
	\caption{The online prediction process: First, the feature map is obtained through model inference, and the features of traversable regions are extracted using recorded odometry data. Then, the prototype vectors are analyzed through online clustering, and finally, the traversability map is generated by computing with the feature map.}
	\label{Inference}
\end{figure}

Off-road environmental scenes are complex, with multiple feasible road surfaces, making it difficult to calculate a traversable vector based on experience for traversability analysis. Therefore, we propose an environment traversability analysis method that integrates real-time driving experiences of the vehicle, as shown in Fig. \ref{Inference}. Specifically, when the autonomous vehicle starts operating, it samples the traversed areas and updates the prototype vector queue \( V_{\text{trav}} = \{ v_i \}_{i=1}^{N} \). Using the Chinese Restaurant Process (CPR), the first sample is assigned to the first prototype vector \( v_1 \). Subsequent samples \( z \) calculate their similarity with each prototype vector \( v_i \) in the queue \( V_{\text{trav}} \), and the highest similarity is identified. The feature vector \( v_{\max} \) corresponding to the maximum similarity is computed as:
\begin{equation}
	v_{max}=\max \left( \text{cosine\_similarity}\left( z,v_i \right) _{i=1}^{N} \right)
\end{equation}

If the similarity is below a predefined threshold \( \alpha \), the sample is considered a new prototype vector and added to the queue \( V_{\text{trav}} \). If the similarity is above \( \alpha \), it indicates that the current environment is consistent with the previously traversed environment, and the corresponding prototype vector \( v_{\max} \) is updated to better fit the current observation of the environment type. The update is performed as follows:
\begin{equation}
	v_\text{new}=m\cdot v_{max}+\left( 1-m \right) \cdot z,
\end{equation}
where \( v_{\text{new}} \) is the updated prototype vector, and \( m \) is the momentum value. Finally, the traversability result $\textbf{T}\in \mathbb{R}^{H\times W}$ is obtained by computing the similarity between the feature map $F$ and the prototype vector queue $V_\text{trav}$, with a range of $[0,1]$. Higher values indicate that the current environment is more similar to previously traversed regions, implying a higher probability of traversability. The probabilistic environmental representation facilitates seamless integration with various planning paradigms in the form of cost maps.
\section{EXPERIMENTS}
\label{EXPERIMENTS}
In this section, we validate the effectiveness of our proposed self-supervised traversability analysis method on different datasets. We first introduce the datasets used for evaluation, followed by the experimental setup and implementation details. Then, we introduce the evaluation metrics and quantitative results. Finally, we present the results of autonomous driving experiments using this method, demonstrating its effectiveness in off-road environments and its suitability for downstream motion planning tasks.
\subsection{Datasets}
\begin{enumerate}
	\item{\textit{Self-collected datasets:} We collected driving data in various environments using a platform equipped with an RGB camera and RS-Ruby LiDAR, accumulating a total driving distance of 30.88 kilometers. The dataset contains 48,600 time-synchronized RGB-point cloud data pairs, covering diverse environmental conditions including urban roads, rural roads, daytime, nighttime, and different seasonal scenarios. To evaluate algorithm performance, we performed fine-grained annotations on the BEV of the test set to establish benchmark ground truth.}
	\item{\textit{RELLIS-3D:} To further validate the universality of the method, we adopted the publicly available RELLIS-3D off-road dataset, which provides RGB images, point clouds, and precise vehicle pose information. Based on the multimodal data provided by the dataset, we constructed a BEV representation of the environment and automatically generated labels using our algorithm. At the same time, the front-view annotations from RELLIS-3D were projected into the BEV space using precise coordinate transformations to create a dependable ground truth. Although the annotation does not indicate which points are traversable, we define the grass, dirt, asphalt, concrete, and mud classes as traversable and the tree, pole, vehicle, object, person, fence, barrier, rubble, and bush classes as non-traversable.}
\end{enumerate}	
\subsection{Experimental Setup}
To comprehensively evaluate the performance of the method, we conducted experiments on both our self-built dataset and RELLIS-3D dataset. $60\%$ of each dataset was used for training, and the remaining $40\%$ was used for testing. To ensure the fairness of the comparative experiments, we reproduced the reconstruction-based autoencoder anomaly detection algorithm\cite{schmid2022self} and the SAM mask-based self-supervised learning algorithm\cite{jung2024v} under the same BEV map input conditions.  The experimental results show that, compared to existing self-supervised traversability analysis algorithms, our method demonstrates significant advantages across all evaluation metrics. Additionally, to validate the effectiveness of the proposed loss function, we designed systematic ablation experiments, proving the efficacy of the loss combination method proposed in this paper.
\subsection{Implementaion Details}
The input to the model is a vehicle-centered BEV of the environment, with the BEV map size being $300\times300$ and a resolution of $0.2$ meter per pixel. The network architecture employs ResNet-34 as the encoder to extract deep feature representations from the BEV map, and a symmetric decoder network is designed to restore the feature maps to the original input resolution through layer-wise upsampling. The model training consists of $60$ epochs with a batch size of $4$, using the Adam optimizer for the optimization process, an initial learning rate of $1 \times 10^{-4}$, and a polynomial learning rate decay strategy. We set the temperature $\tau=0.05$, the weight $\lambda = \min(1,\text{epoch}/60)$, the momentum value $m=0.99$ and the update threshold $\alpha=0.9$. Employing a multi-level clustering approach, cluster centers are established at $50$, $100$, and $500$ to capture hierarchical feature representations. 
\begin{figure*}
	\centering
	\begin{spacing}{0.15} 
		\setlength\tabcolsep{2pt} 
		\begin{tabular}{ccccccc}
			& \textbf{Front View} & \textbf{Input BEV} & \textbf{Ground Truth} & \textbf{Schmid \textit{et al.}} & \textbf{Jung \textit{et al.}} & \textbf{Ours}  \\
			\\
			\textbf{Paved} &  
			\begin{minipage}[b]{0.2\linewidth}
				\centering
				\raisebox{-.5\height}{\includegraphics[width=\linewidth]{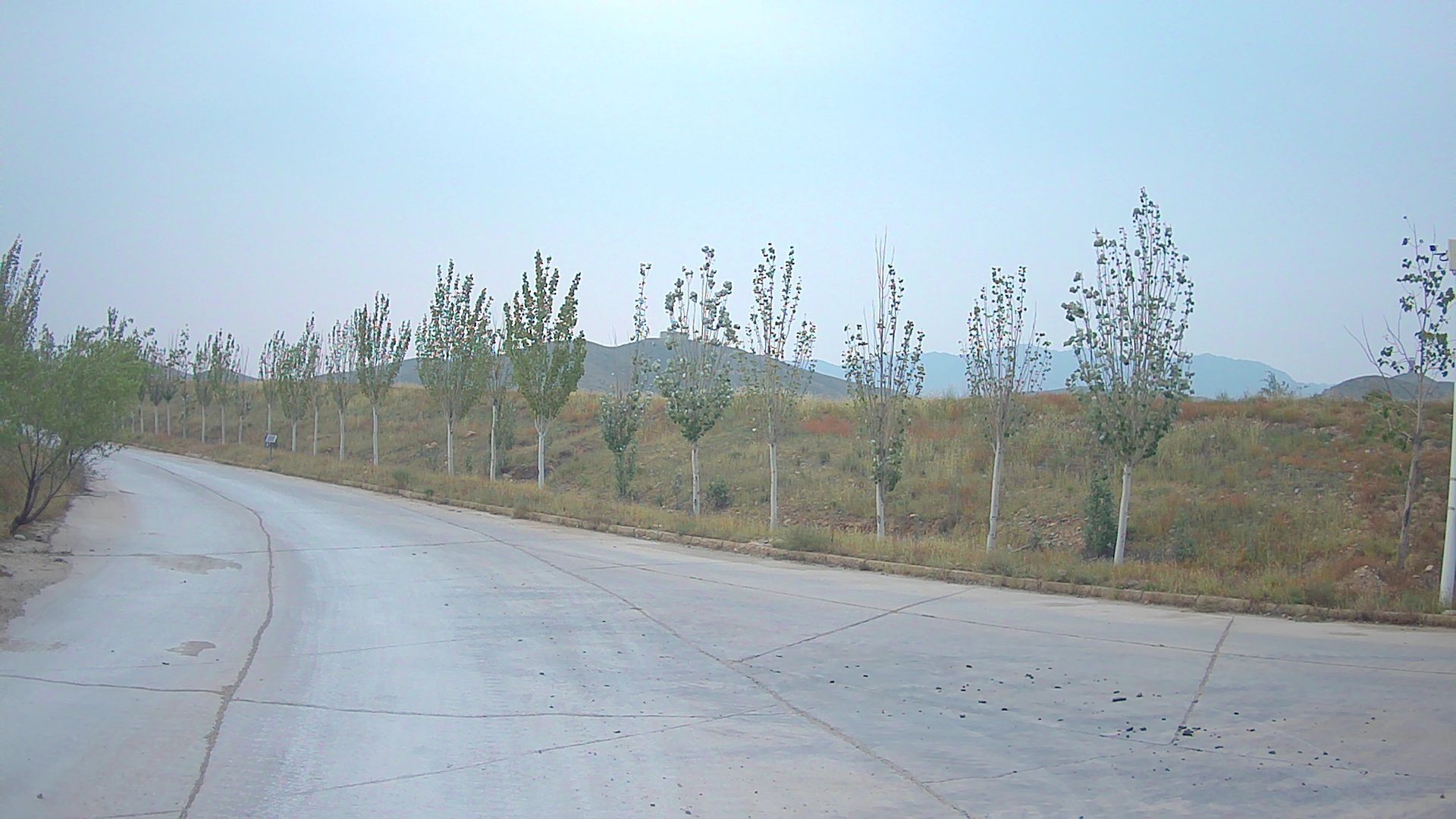}}
			\end{minipage} &  
			\begin{minipage}[b]{0.112\linewidth}
				\centering
				\raisebox{-.5\height}{\includegraphics[width=\linewidth]{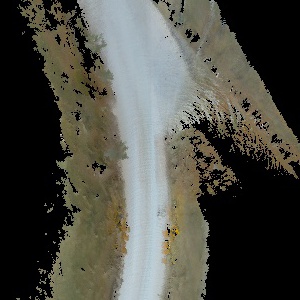}}
			\end{minipage} &     
			\begin{minipage}[b]{0.112\linewidth}
				\centering
				\raisebox{-.5\height}{\includegraphics[width=\linewidth]{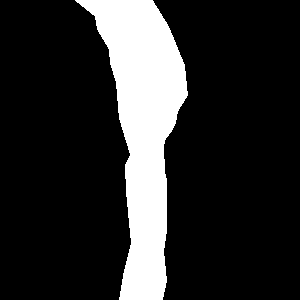}}
			\end{minipage} &    
			\begin{minipage}[b]{0.112\linewidth}
				\centering
				\raisebox{-.5\height}{\includegraphics[width=\linewidth]{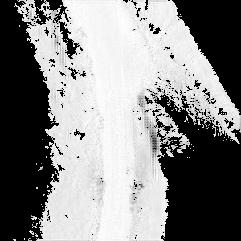}}
			\end{minipage} &   
			\begin{minipage}[b]{0.112\linewidth}
				\centering
				\raisebox{-.5\height}{\includegraphics[width=\linewidth]{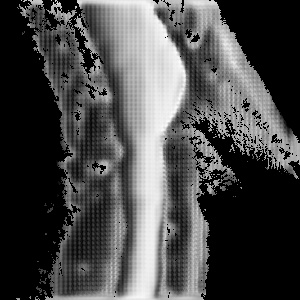}}
			\end{minipage} &      
			\begin{minipage}[b]{0.112\linewidth}
				\centering
				\raisebox{-.5\height}{\includegraphics[width=\linewidth]{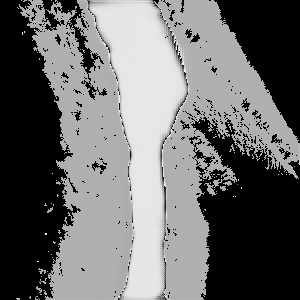}}
			\end{minipage}              
			\\
			\\ 
			\textbf{Mountain Road} &
			\begin{minipage}[b]{0.2\linewidth}
				\centering
				\raisebox{-.5\height}{\includegraphics[width=\linewidth]{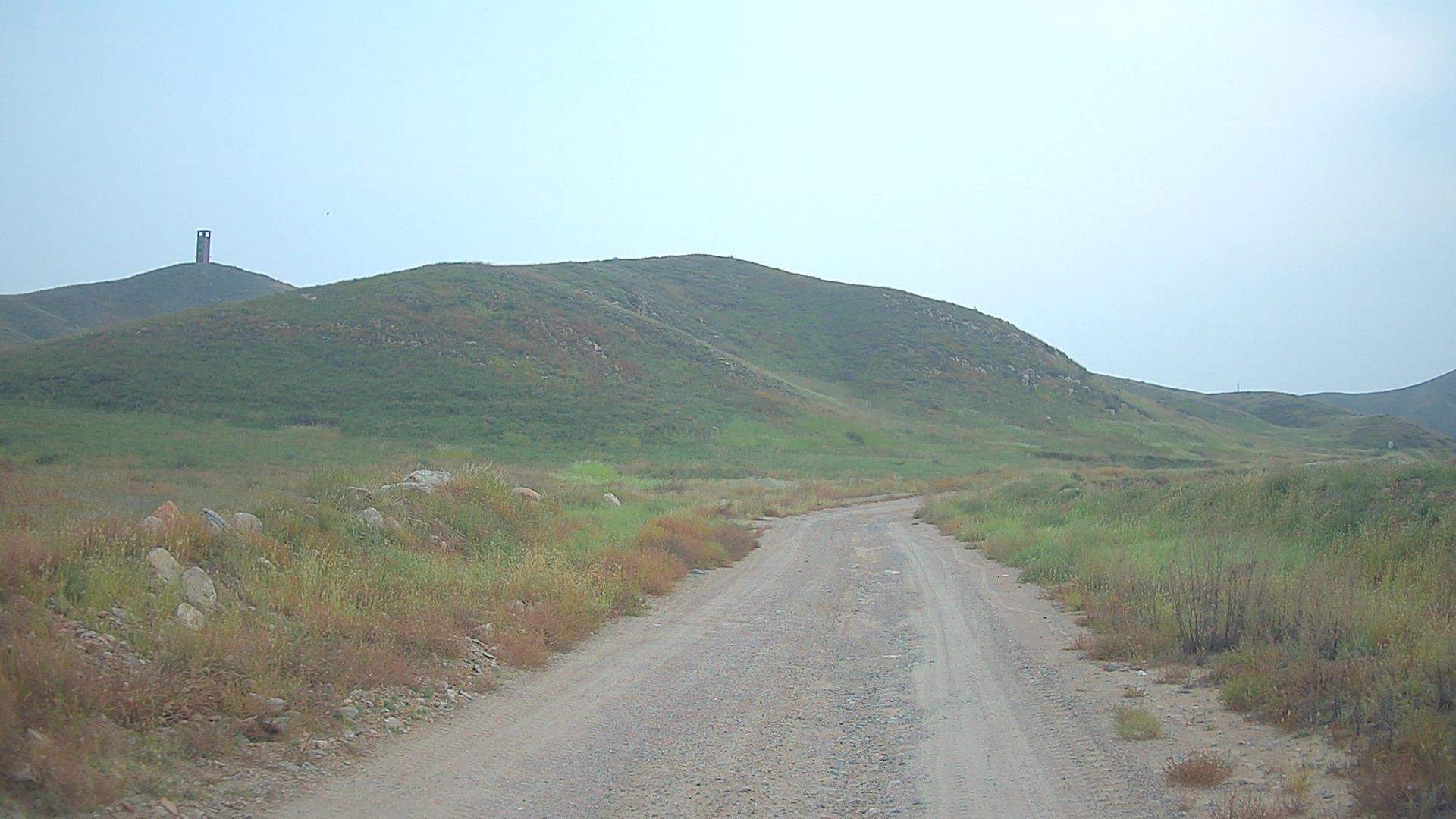}}
			\end{minipage} &    
			\begin{minipage}[b]{0.112\linewidth}
				\centering
				\raisebox{-.5\height}{\includegraphics[width=\linewidth]{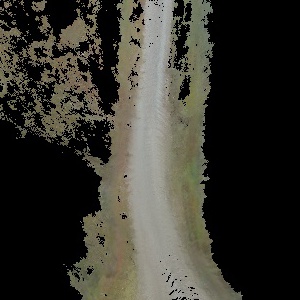}}
			\end{minipage} &     
			\begin{minipage}[b]{0.112\linewidth}
				\centering
				\raisebox{-.5\height}{\includegraphics[width=\linewidth]{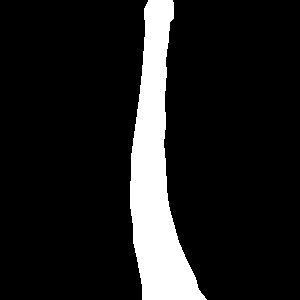}}
			\end{minipage} &
			\begin{minipage}[b]{0.112\linewidth}
				\centering
				\raisebox{-.5\height}{\includegraphics[width=\linewidth]{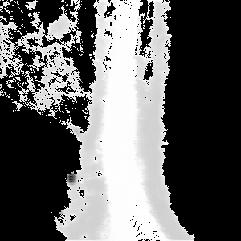}}
			\end{minipage} &     
			\begin{minipage}[b]{0.112\linewidth}
				\centering
				\raisebox{-.5\height}{\includegraphics[width=\linewidth]{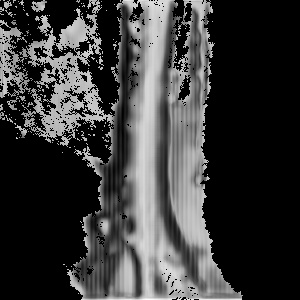}}
			\end{minipage} &      
			\begin{minipage}[b]{0.112\linewidth}
				\centering
				\raisebox{-.5\height}{\includegraphics[width=\linewidth]{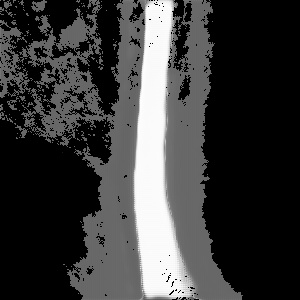}}
			\end{minipage}              
			\\
			\\ 
			\textbf{Spring/Summer} &  
			\begin{minipage}[b]{0.2\linewidth}
				\centering
				\raisebox{-.5\height}{\includegraphics[width=\linewidth]{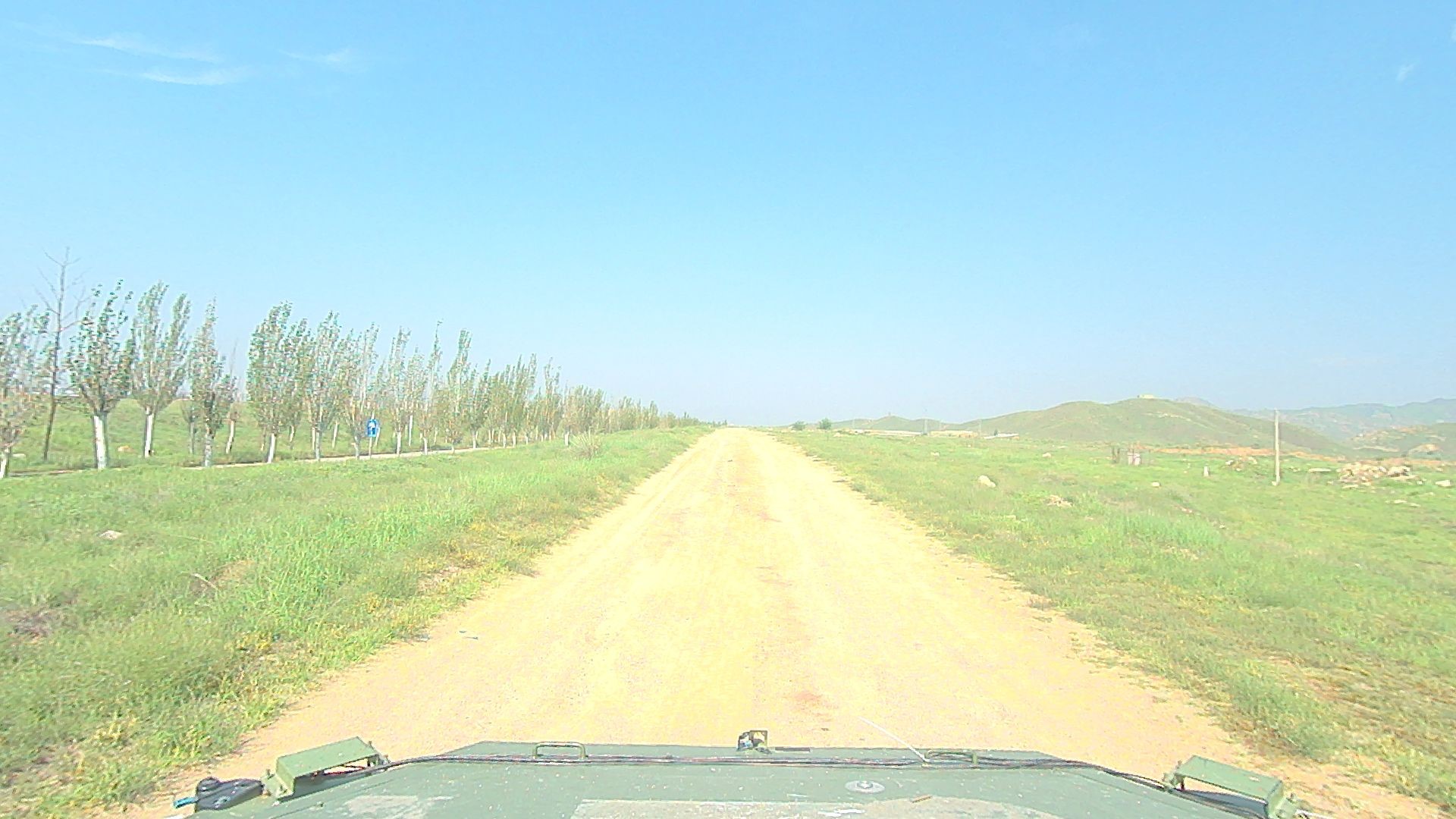}}
			\end{minipage} &   
			\begin{minipage}[b]{0.112\linewidth}
				\centering
				\raisebox{-.5\height}{\includegraphics[width=\linewidth]{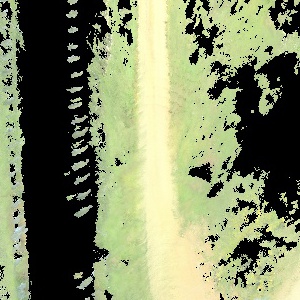}}
			\end{minipage} &     
			\begin{minipage}[b]{0.112\linewidth}
				\centering
				\raisebox{-.5\height}{\includegraphics[width=\linewidth]{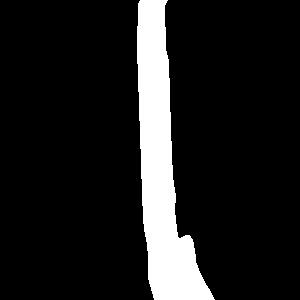}}
			\end{minipage} &    			
			\begin{minipage}[b]{0.112\linewidth}
				\centering
				\raisebox{-.5\height}{\includegraphics[width=\linewidth]{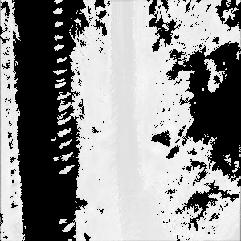}}
			\end{minipage} &    
			\begin{minipage}[b]{0.112\linewidth}
				\centering
				\raisebox{-.5\height}{\includegraphics[width=\linewidth]{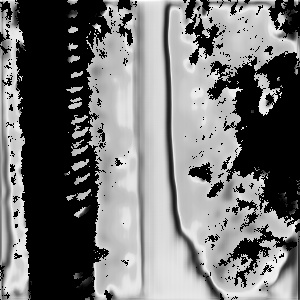}}
			\end{minipage} &      
			\begin{minipage}[b]{0.112\linewidth}
				\centering
				\raisebox{-.5\height}{\includegraphics[width=\linewidth]{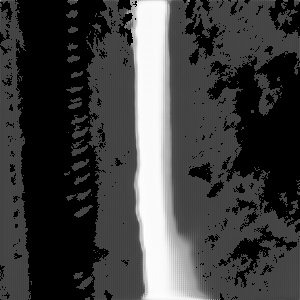}}
			\end{minipage}              \\ 
			\\
			\textbf{Autumn/Winter} &  
			\begin{minipage}[b]{0.2\linewidth}
				\centering
				\raisebox{-.5\height}{\includegraphics[width=\linewidth]{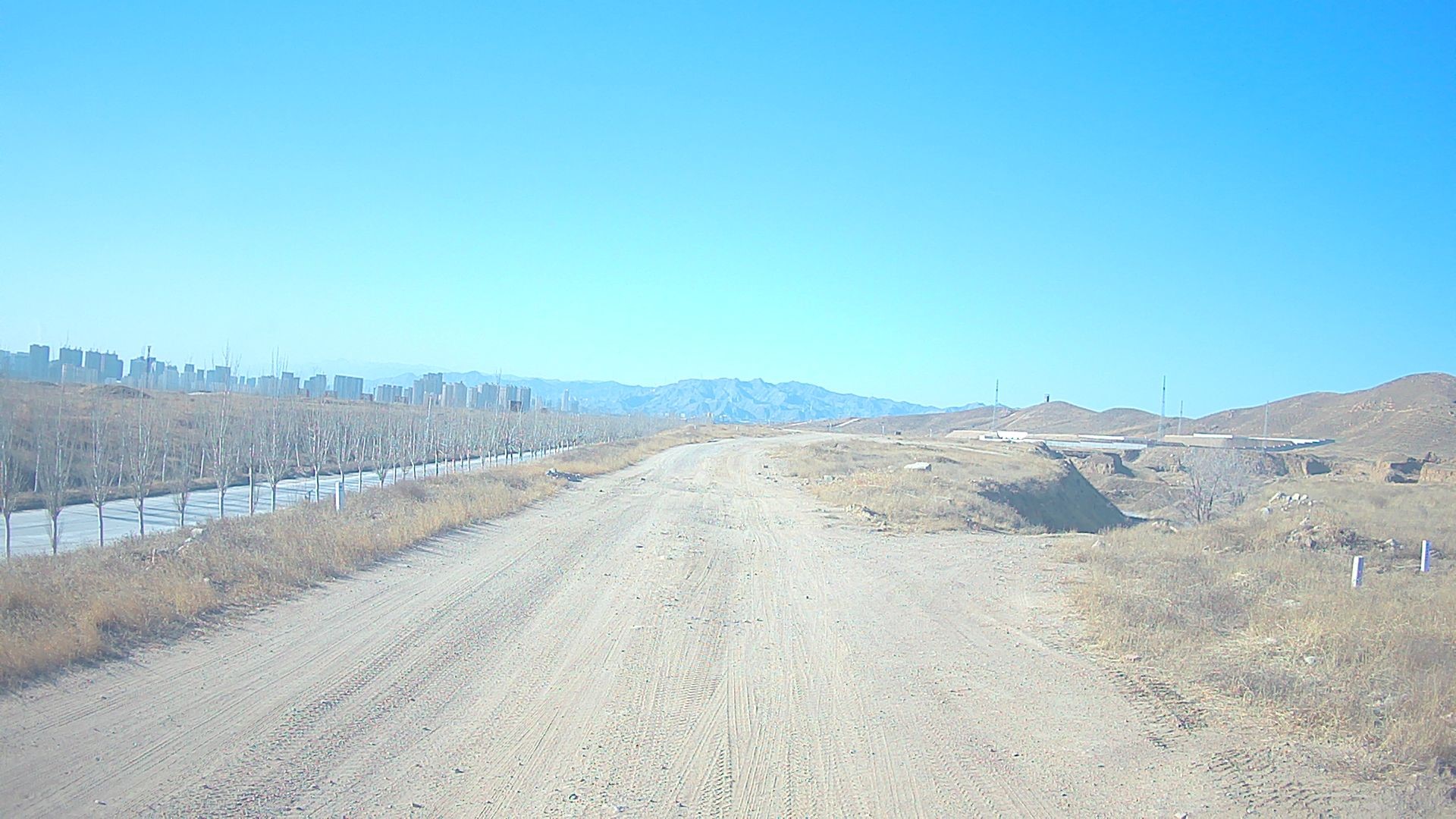}}
			\end{minipage} &  
			\begin{minipage}[b]{0.112\linewidth}
				\centering
				\raisebox{-.5\height}{\includegraphics[width=\linewidth]{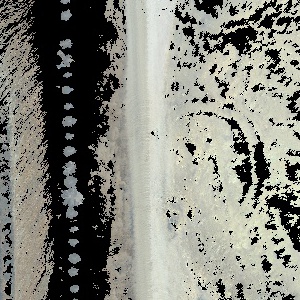}}
			\end{minipage} &     
			\begin{minipage}[b]{0.112\linewidth}
				\centering
				\raisebox{-.5\height}{\includegraphics[width=\linewidth]{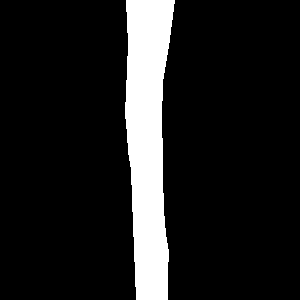}}
			\end{minipage} &     			
			\begin{minipage}[b]{0.112\linewidth}
				\centering
				\raisebox{-.5\height}{\includegraphics[width=\linewidth]{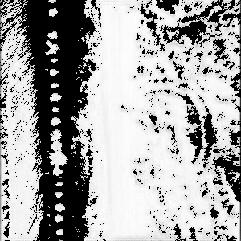}}
			\end{minipage} & 
			\begin{minipage}[b]{0.112\linewidth}
				\centering
				\raisebox{-.5\height}{\includegraphics[width=\linewidth]{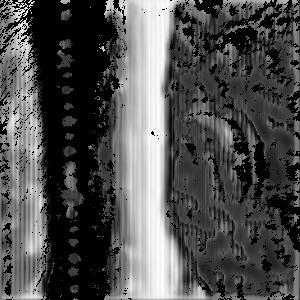}}
			\end{minipage} &      
			\begin{minipage}[b]{0.112\linewidth}
				\centering
				\raisebox{-.5\height}{\includegraphics[width=\linewidth]{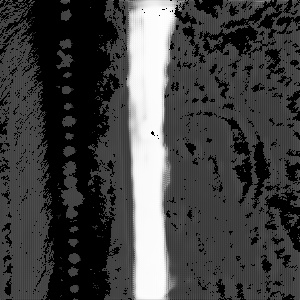}}
			\end{minipage}    \\
			
			\\
			\textbf{RELLIS-3D} &  
			\begin{minipage}[b]{0.2\linewidth}
				\centering
				\raisebox{-.5\height}{\includegraphics[width=\linewidth]{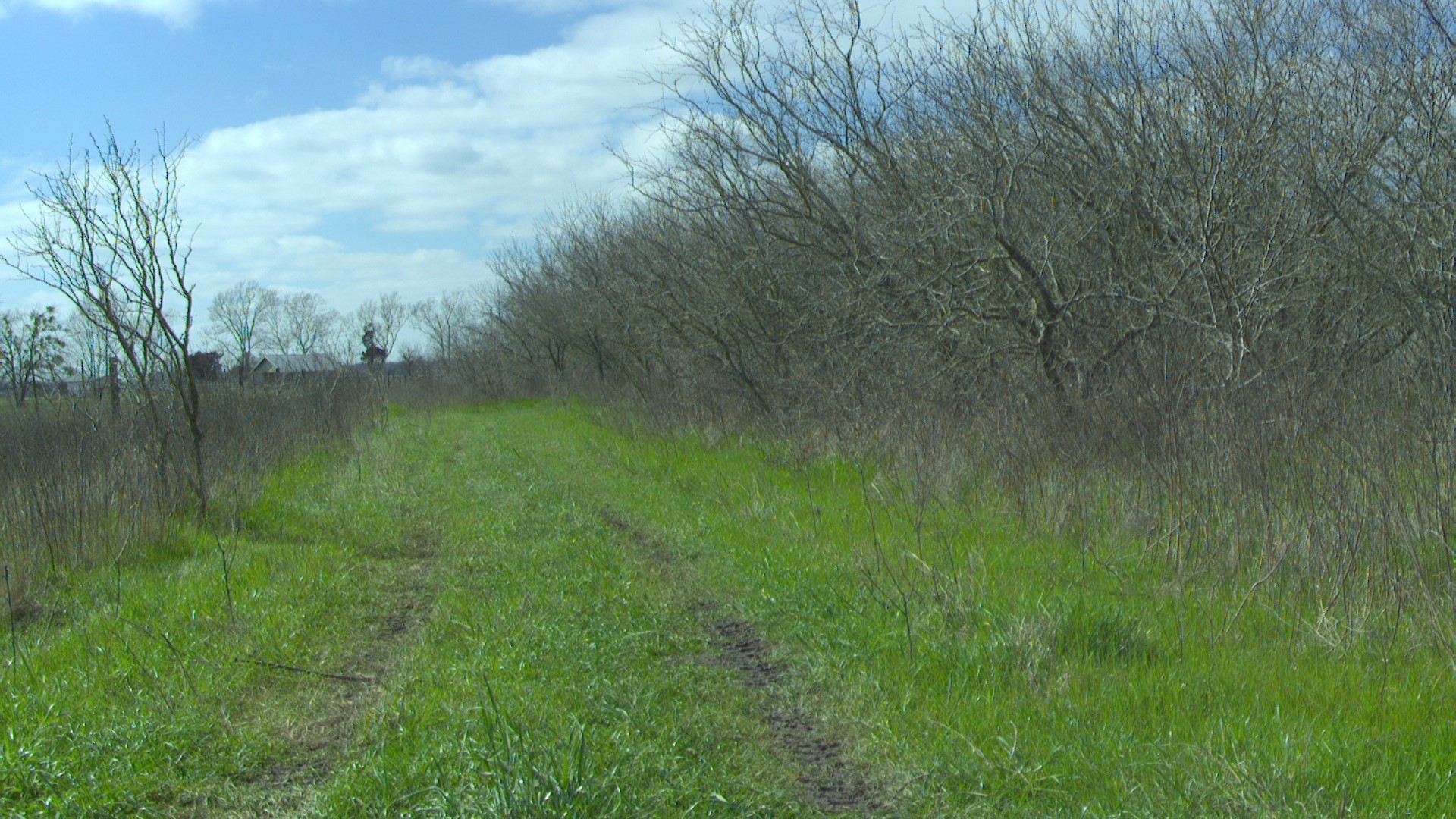}}
			\end{minipage} & 
			\begin{minipage}[b]{0.112\linewidth}
				\centering
				\raisebox{-.5\height}{\includegraphics[width=\linewidth]{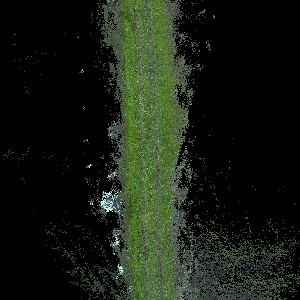}}
			\end{minipage} &     
			\begin{minipage}[b]{0.112\linewidth}
				\centering
				\raisebox{-.5\height}{\includegraphics[width=\linewidth]{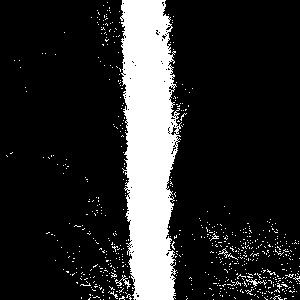}}
			\end{minipage} &     
			\begin{minipage}[b]{0.112\linewidth}
				\centering
				\raisebox{-.5\height}{\includegraphics[width=\linewidth]{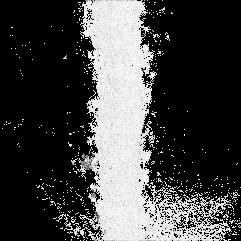}}
			\end{minipage} &  
			\begin{minipage}[b]{0.112\linewidth}
				\centering
				\raisebox{-.5\height}{\includegraphics[width=\linewidth]{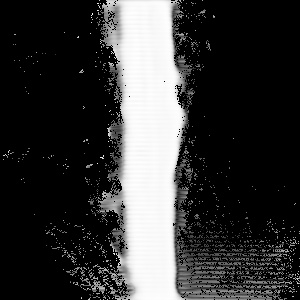}}
			\end{minipage} &      
			\begin{minipage}[b]{0.112\linewidth}
				\centering
				\raisebox{-.5\height}{\includegraphics[width=\linewidth]{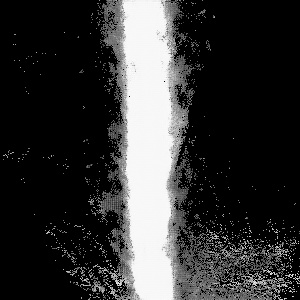}}
			\end{minipage} 
		\end{tabular}
	\end{spacing}
	\vspace{2em}
	\caption{The input BEV, ground truth, qualitative results from Schmid et al.\cite{schmid2022self} , Jung et al.\cite{jung2024v} , and our method on both our self-collected dataset and the RELLIS-3D dataset. We employ a probabilistic representation with a range of $[0,1]$, where darker regions indicate lower traversability probabilities and brighter regions correspond to higher traversability probabilities.}
	\label{Result}
\end{figure*}
\subsection{Experimental Results}
The quantitative results on our dataset and the RELLIS-3D dataset is shown in Tab.\ref{tab:performance}. The datasets listed from top to bottom are paved roads, mountain roads, spring/summer off-road scenarios, autumn/winter off-road scenarios, and RELLIS-3D. To achieve quantitative evaluation, the continuous traversability probability estimation is converted into binary classification results by setting a decision threshold $\tau$. Evaluation metrics include: Area Under the Receiver Operating Characteristic curve (AUROC), F1 score, Average Precision (AP), along with Precision (PRE), Recall (REC), False Positive Rate (FPR), and False Negative Rate (FNR) computed at the optimal threshold $\tau^*$ determined by maximizing the F1 score. The qualitative results on our dataset and the RELLIS-3D dataset is shown in Fig. \ref{Result}. Compared to other self-supervised methods, our approach demonstrates significant advantages. Notably, reconstruction-based autoencoder anomaly detection algorithms tend to produce a high number of false positives. The generalization capability of autoencoders makes it difficult for anomaly detection methods to effectively distinguish non-traversable regions in traversability analysis. SAM mask-based self-supervised learning algorithms struggle to accurately distinguish road regions in top-down views, leading to many false positives and false negatives during training, which adversely affects the differentiation between traversable and non-traversable areas. Additionally, in real off-road scenarios, traversable surfaces consist of various materials, and relying on a single, fixed traversability vector obtained during training is insufficient for long-duration autonomous driving.

This method integrates multiple sensor data. During training, a reliable obstacle detection algorithm is used to precisely mark some of the non-traversable regions in the environment, and driving data obtained from manual driving is used as positive samples to provide accurate positive and negative samples.  Furthermore, pseudo-labeling techniques make full use of the large amount of unlabeled data in the samples, promoting more efficient and faster convergence of the network. During inference, real-time vehicle experience is used for online clustering, continuously updating traversability vectors to achieve adaptive adjustments, thereby improving the accuracy of traversability analysis and the generalization ability of the algorithm. Additionally, using BEV for environment traversability analysis can effectively integrate key environmental factors related to autonomous driving (such as road topology and obstacle distribution), reduce irrelevant input, and thus reduce the waste of computational resources.
\begin{table}[!t]
	\begin{center}
		\caption{Performance comparison of different methods.}
		\label{tab:performance}
		\tiny
		\begin{tabular}{l|c|c|c|c|c|c|c}
			\toprule
			\textbf{Methods} & \textbf{AUROC} $\uparrow$ & \textbf{AP} $\uparrow$ & \textbf{F1} $\uparrow$ & \textbf{FPR} $\downarrow$ & \textbf{FNR} $\downarrow$ & \textbf{Pre.} $\uparrow$ & \textbf{Rec.} $\uparrow$ \\
			\midrule
			Schmid \textit{et al.}\cite{schmid2022self} & 0.092 & 0.179 & 0.756 & 0.068 & 0.290 & 0.817 & 0.710 \\
			Jung \textit{et al.}\cite{jung2024v} & 0.895 & 0.867 & 0.724 & 0.275 & \textbf{0.164} & 0.666 & \textbf{0.836} \\
			\textbf{Ours} & \textbf{0.919 }& \textbf{0.937 }& \textbf{0.888} &\textbf{ 0.005} & 0.183 & \textbf{0.988} & 0.817 \\
			\midrule
			Schmid \textit{et al.}\cite{schmid2022self} & 0.036 & 0.159 & 0.729 & 0.284 & \textbf{0.007} & 0.578 & \textbf{0.993} \\
			Jung \textit{et al.}\cite{jung2024v} & 0.963 & 0.934 & 0.846 & 0.099 & 0.112 & 0.820 & 0.888 \\
			\textbf{Ours} & \textbf{0.984} & \textbf{0.977} & \textbf{0.930} & \textbf{0.0132} & 0.100 & \textbf{0.963} & 0.900 \\
			\midrule
			Schmid \textit{et al.}\cite{schmid2022self} & 0.331 & 0.218 & 0.468 & 0.174 & 0.425 & 0.398 & 0.575 \\
			Jung \textit{et al.} & 0.937 & 0.890 & 0.824 & 0.076 & 0.151 & 0.814 & 0.849 \\
			\textbf{Ours} & \textbf{0.987} & \textbf{0.924} & \textbf{0.894} & \textbf{0.039 }& \textbf{0.059} &\textbf{ 0.857} & \textbf{0.941} \\
			\midrule
			Schmid \textit{et al.}\cite{schmid2022self} & 0.205 & 0.099 & 0.347 & 0.684 & 0.054 & 0.2132 & 0.946 \\
			Jung \textit{et al.}\cite{jung2024v} & 0.966 & 0.877 & 0.638 & 0.223 &\textbf{ 0.043} & 0.489 & \textbf{0.957} \\
			\textbf{Ours} & \textbf{0.977} & \textbf{0.954} & \textbf{0.902} & \textbf{0.014} & 0.111 & \textbf{0.924} & 0.889 \\
			\midrule
			Schmid \textit{et al.}\cite{schmid2022self} & 0.443 & 0.518 & 0.687 & 0.808 & 0.132 & 0.576 & 0.862 \\
			Jung \textit{et al.}\cite{jung2024v} & 0.842 & 0.838 & 0.794 & 0.564 & \textbf{0.047} & 0.683 & \textbf{0.953} \\
			\textbf{Ours} & \textbf{0.936} & \textbf{0.945} & \textbf{0.890} & \textbf{0.194} &  0.081 & \textbf{0.867} &  0.919 \\
			\bottomrule
		\end{tabular}
	\end{center}
\end{table}
\begin{figure}[!h]
	\centering
	\subfloat{
		\centering
		\includegraphics[width = 0.27\linewidth]{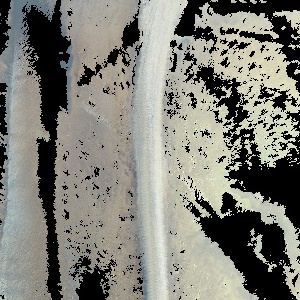}
		\includegraphics[width = 0.27\linewidth]{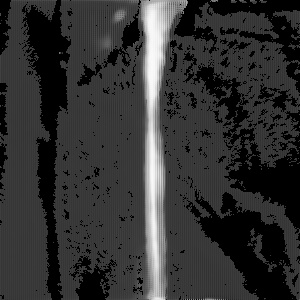}
		\includegraphics[width = 0.27\linewidth]{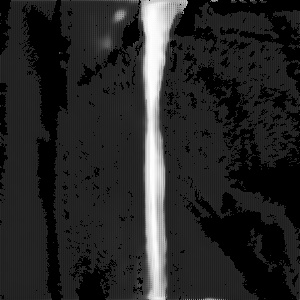}
	}
	\caption{(Left) The input BEV. (Middle) The traversability results use fixed prototype vectors. (Right) The traversability results use prototype vectors adaptively computed online.}
	\label{fig:Ablation}
\end{figure}
\begin{figure}[h]
	\centering
	\subfloat{
		\centering
		\includegraphics[width = 0.42\linewidth]{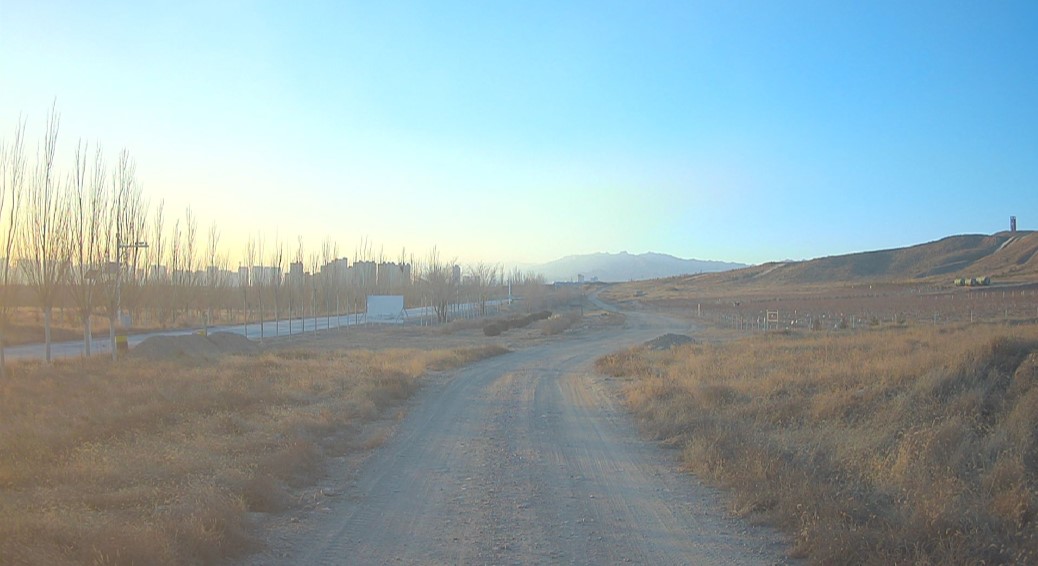}
		\includegraphics[width = 0.23\linewidth]{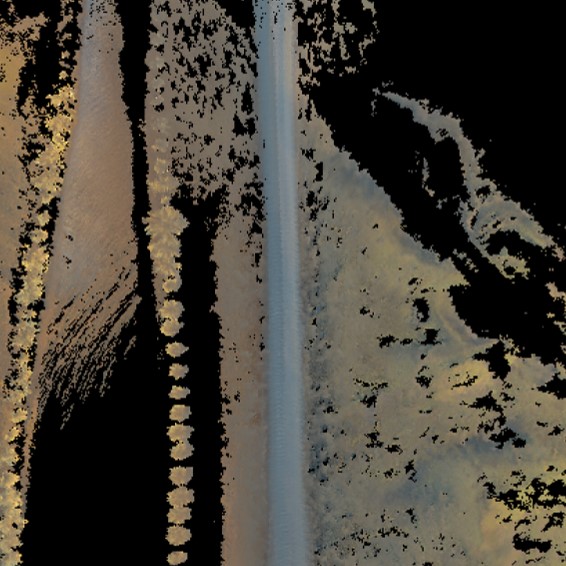}
		\includegraphics[width = 0.23\linewidth]{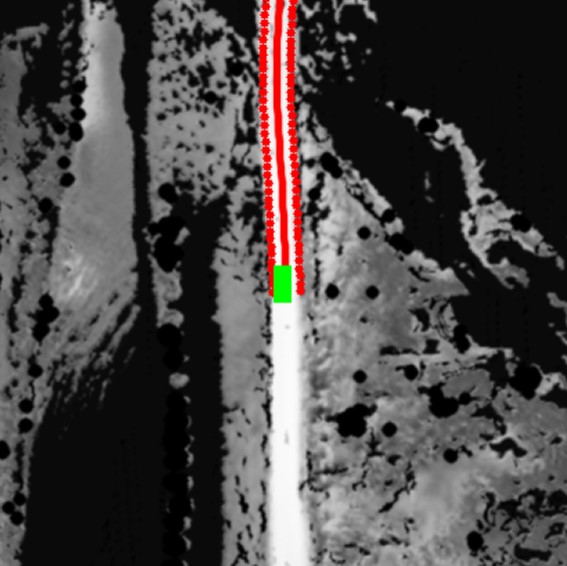}
	}
	\caption{(Left) RGB images captured by the onboard front-facing camera. (Middle) The input bev. (Right) Results of motion planning based on traversability analysis.}
	\label{fig:plan}
\end{figure}
\subsection{Ablation Study}
This study employs a multi-level feature clustering mechanism, optimizing the feature space at multiple granularities to ensure that samples with similar attributes are more compactly distributed in the feature space, thereby enabling prototype vectors to infer traversability more effectively. To verify the effectiveness of each component of our method, we designed a series of ablation experiments. These experiments first compare the training performance of different loss function combinations and then analyze the performance differences between fixed prototype vectors and online adaptive prototype vectors updated based on real-time vehicle operation data. The results are shown in Tab.\ref{tab:ablation}. For simplicity, we abbreviate $\mathcal{L}_{contrast}$ as $\mathcal{L}_{1}$, $\mathcal{L}_{cluster}$ as $\mathcal{L}_{2}$, $\mathcal{L}_{unlabel}$ as $\mathcal{L}_{3}$.

Notably, the quantitative and qualitative evaluations of models trained on spring/summer data and tested on autumn/winter datasets (see Tab. \ref{tab:ablation} and Fig. \ref{fig:Ablation}) indicate that online adaptive prototype vectors significantly enhance the model's cross-season generalization ability. The performance improvement is primarily attributed to the online adaptation mechanism for dynamic road condition changes.
\begin{table}[h]
	\begin{center}
		\caption{Quantitative results of the ablation studies.}
		\label{tab:ablation}
		\scriptsize
		\begin{tabular}{l|c|c|c|c|c}
			\toprule
			\quad   & \textbf{AUROC} $\uparrow$ & \textbf{AP} $\uparrow$ & \textbf{F1} $\uparrow$ & \textbf{FPR} $\downarrow$ & \textbf{FNR} $\downarrow$ \\ \midrule
			Solely \textbf{w/ $\mathcal{L}_{1}$}                          &           0.932           &         0.945          &         0.884          &           0.170           &      \textbf{0.107}             \\
			\textbf{w/ $\mathcal{L}_{1}+\mathcal{L}_{2}$}                 &           0.941           &         0.953          &         0.884          &           0.132           &           0.127                     \\
			\textbf{w/ $\mathcal{L}_{1}+\mathcal{L}_{3}$}                 &           0.936           &         0.950          &         0.880          &           0.152           &           0.123                     \\
			\textbf{w/ $\mathcal{L}_{2}+\mathcal{L}_{3}$}                 &           0.709           &         0.770          &         0.588          &           0.167           &           0.517                    \\
			\textbf{w/ $\mathcal{L}_{1}+\mathcal{L}_{2}+\mathcal{L}_{3}$} &      \textbf{0.943}       &     \textbf{0.955}     &     \textbf{0.888}     &      \textbf{0.125}       &           0.127                      \\ \midrule 
			\textbf{Fixed Prototype} &      0.932       &     
			0.862     &     
			0.745 		&      0.345      &           \textbf{0.014}       \\
			\textbf{Adaptive Prototype} &      \textbf{0.962}       &     \textbf{0.898}     &     \textbf{0.807}     &      \textbf{0.194}       &           0.051       \\
			\bottomrule
		\end{tabular}
	\end{center}
\end{table}
\subsection{Field Validation Through Vehicle Testing}
This method achieves real-time processing performance at 10Hz on the NVIDIA L4 computing platform. The BEV representation aligns naturally with the spatial coordinate system of motion planning, making it more suitable for downstream motion planning. As shown in Fig. \ref{fig:plan}, the traversability analysis system based on this method successfully completed 5.5km of continuous autonomous navigation in complex off-road scenarios, with an average speed of 27 km/h.

\section{CONCLUSIONS}
\label{CONCLUSIONS}
This paper proposes a novel self-supervised method that can learn the traversability of the environment without any manual annotations. In the online process, real-time vehicle experiences are used to analyze traversability, and a probabilistic environment representation is adopted, which is more suitable for downstream motion planning tasks. Extensive experiments show that the method proposed in this paper is more efficient than other methods in learning traversability. However, the current work assumes that the vehicle always drives within the traversable area, which introduces some subjectivity.

In future work, we plan to combine the geometric information of the environment with the vehicle's own perception information to perform a more objective analysis of traversability and obtain a more precise traversability description.



%
%
%
%
%
%
%

\bibliographystyle{IEEEtran}
\bibliography{root}
\end{document}